\documentclass[journal]{IEEEtran}

\usepackage{times}
\usepackage{soul}
\usepackage{url}
\usepackage{hyperref}
\usepackage[utf8]{inputenc}
\usepackage[small]{caption}
\usepackage{amsfonts,amssymb}
\usepackage{graphicx}
\usepackage{amsmath}
\usepackage{booktabs}
\usepackage{algorithm}
\usepackage{algorithmic}
\usepackage{multirow}
\usepackage{xspace}
\usepackage{subfigure}
\usepackage{footnote}
\usepackage{color}
\usepackage{wasysym}
\usepackage{bm}

\let\oldhat\hat
\renewcommand{\vec}[1]{\mathbf{#1}}
\renewcommand{\hat}[1]{\oldhat{\mathbf{#1}}}
\renewcommand{\matrix}[1]{\mathbf{#1}}

\usepackage{epstopdf}
\usepackage{threeparttable}
\usepackage{dcolumn}
\newcolumntype{d}[1]{D{.}{.}{#1}}

\newcommand{\eg}{\emph{e.g.,}\xspace}
\newcommand{\rf}{\emph{rf.}\xspace}

\newcommand{\ie}{\emph{i.e.,}\xspace}
\newcommand{\etc}{\emph{etc.}\xspace}
\newcommand{\etal}{\emph{et al.}\xspace}
\newcommand{\aka}{\emph{a.k.a.,}\xspace}

\newcommand{\eat}[1]{}
\newcommand{\paratitle}[1]{\vspace{1ex}\noindent \textbf{#1}}
\begin{document}

\title{Exploiting Global Contextual Information for Document-level Named Entity Recognition}

\author{Zanbo Wang, Wei Wei\IEEEauthorrefmark{1}
, Xianling Mao, Shanshan Feng, Pan Zhou, Zhiyong He, Sheng Jiang

\thanks{This work was supported in part by the National Natural Science Foundation of China under Grant No. 61602197 and Grant No. 61772076, Grant No. 61972448, Grant No. L1924068 and in part by Equipment Pre-Research Fund for The 13th Five-year Plan under Grant No. 41412050801.}
\thanks{*Corresponding author, E-mail addresses: weiw@hust.edu.cn (W. Wei)}
\thanks{Z. Wang, W. Wei and S. Jiang are with the School of Computer Science and Technology, Huazhong University of Science and Technology.}
\thanks{X. Mao is with the School of Computer, Beijing Institute of Technology.}
\thanks{S. Feng is with the Inception Institute of Artificial Intelligence Abu Dhabi, UAE.}
\thanks{P. Zhou is with the School of Cyber Science and Engineering, Huazhong University of Science and Technology.}
\thanks{Z. He is with the School of Electronic Engineering, Naval University of Engineering.}
}

%
%


\maketitle

\begin{abstract}
Most existing \emph{named entity recognition} (NER) approaches are based on sequence labeling models, which focus on capturing the local context dependencies.
However, the way of taking one sentence as input prevents the modeling of non-sequential global context, which is useful especially when local context information is limited or ambiguous.
To this end, we propose a model called \underline{\textbf{G}}lobal \underline{\textbf{C}}ontext enhanced \underline{\textbf{Doc}}ument-level NER (\textsf{GCDoc})
to leverage global contextual information from two levels, \ie both word and sentence.
At word-level,
a document graph is constructed to model a wider range of dependencies between words,
then obtain an enriched contextual representation for each word via \emph{graph neural networks} (GNN).
To avoid the interference of noise information, we further propose two strategies.
First we apply the epistemic uncertainty theory to find out tokens whose representations are less reliable,
thereby helping prune the document graph.
Then a \emph{selective auxiliary classifier} is proposed to effectively learn the weight of edges in document graph
and reduce the importance of noisy neighbour nodes.
At sentence-level, for appropriately modeling wider context beyond single sentence,
we employ a cross-sentence module which encodes adjacent sentences and fuses it with the current sentence representation
via attention and gating mechanisms.
Extensive experiments on two benchmark NER datasets (CoNLL 2003 and Ontonotes 5.0 English dataset) demonstrate the effectiveness of our proposed model.
Our model reaches $F_1$ score of 92.22$\pm$0.02 (93.40$\pm$0.02 with BERT) on CoNLL 2003 dataset and 88.32$\pm$0.04 (90.49$\pm$0.09 with BERT) on Ontonotes 5.0 dataset,
achieving new state-of-the-art performance.


\end{abstract}


\IEEEpeerreviewmaketitle

\section{Introduction}

Named Entity Recognition (\aka entity chunking),
which aims to identify and classify
text spans that mention named entities
into the pre-defined categories, \eg persons, organizations and locations, \etc,
is one of the fundamental sub-tasks in information extraction (IE).
With advances in deep learning, much research effort~\cite{huang2015bidirectional, Lample2016Neural, luo2019hierarchical} has been dedicated to enhancing NER systems by utilizing \emph{neural networks} (NNs) to automatically extract features, yielding state-of-the-art performance.

Most existing approaches address NER by sequence labeling models~\cite{Ma2016End, chen2019grn}, which process each sentence independently. However, in many scenarios, entities need to be extracted from a document, such as the financial field. The way of taking single sentence as input prevents the utilization of global information from the entire document. Intead of sentence-level NER, we focus on studying the document-level NER problem.


\begin{figure}[!t]
\subfigure[]{
\begin{minipage}[b]{0.51\textwidth}
\includegraphics[width=\textwidth] {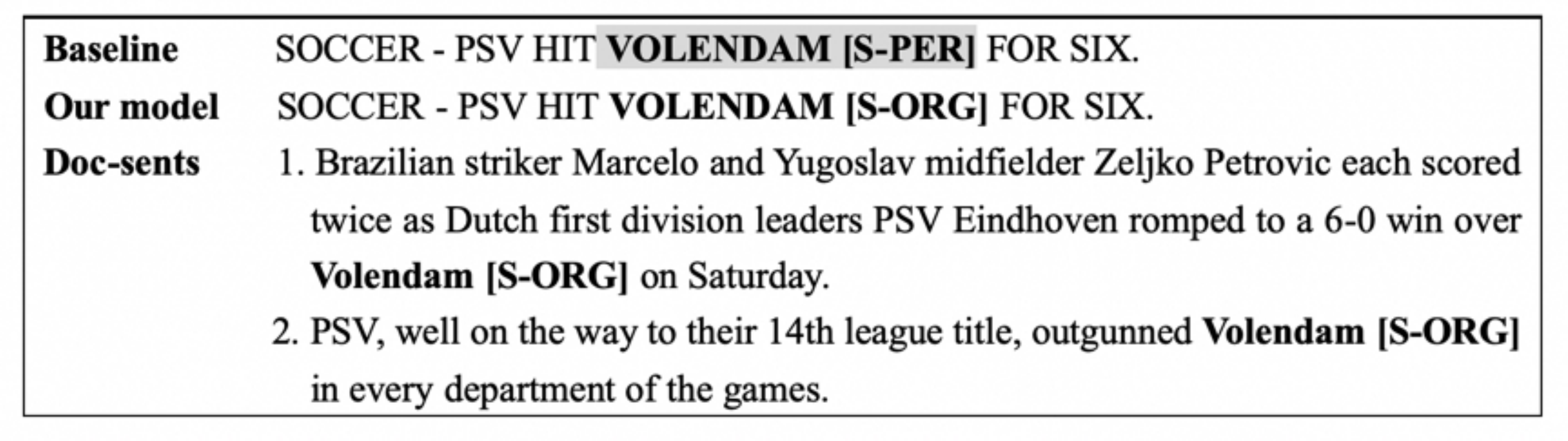}
\end{minipage}
\label{fig:examA}
}
\subfigure[]{
\begin{minipage}[b]{0.51\textwidth}
\includegraphics[width=\textwidth]{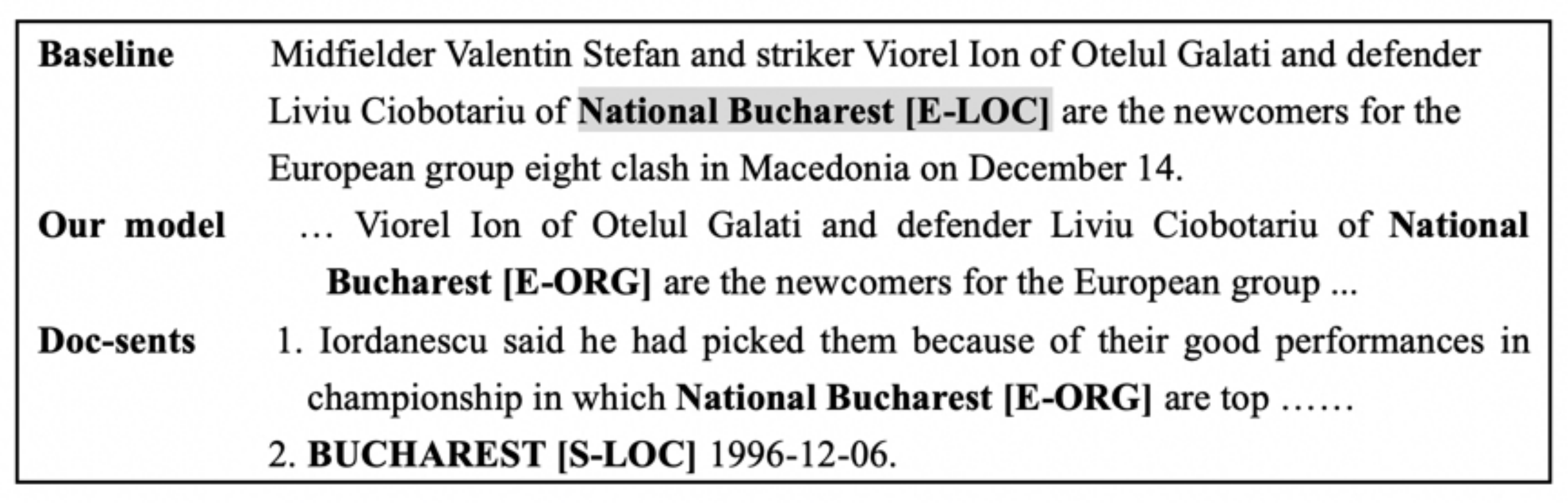}
\end{minipage}
\label{fig:examB}
}
\subfigure[]{
\begin{minipage}[b]{0.51\textwidth}
\includegraphics[width=\textwidth]{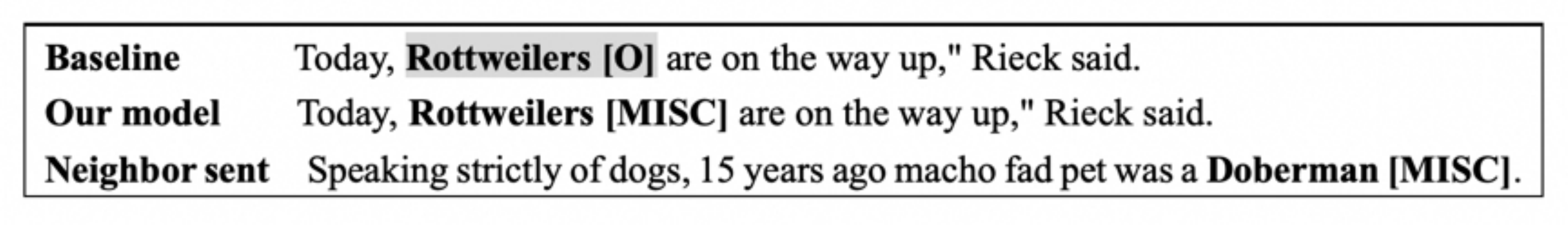}
\end{minipage}
\label{fig:examC}
}
\caption{Examples from the baseline and our model to illustrate our motivation.}
\end{figure}

Named entity mentions appearing repeatedly in a document
may be naturally topically relevant to the current entity
and thus can provide external context for 
entity understanding,
for example, in Figure \ref{fig:examA}, the baseline model mistakenly tags ``VOLENDAM" as a person (PER) instead of organization (ORG) owing to the limited local context, while
some text in the two other sentences within the same document in Figure \ref{fig:examA} such as ``6-0 win over" clearly implies ``VOLENDAM" to represent a team that should be tagged as ORG. 
%
Additionally,
the semantic relevances are naturally distributed
into the adjacent sentences.
For example, in the case of Figure \ref{fig:examC}, the baseline model fails to recognize the \emph{OOV} token ``Rottweilers" as a Miscellaneous (MISC) entity, 
while the previous sentence 
actually has provided an obvious hint that it specifically refers to a Pet name like ``Doberman''.
Therefore, in this paper we propose to fully and effectively exploit
the rich global contextual information for enhancing the performance of document-level NER.

However, the use of extra information will inevitably introduce noise. Since the entity category of words that repeatedly appear also can be different, and adjacent sentences might bring semantic interference.
For example, in Figure \ref{fig:examB}, 
the first sentence provides useful information for tagging ``Bucharest" as ORG in the query sentence, whereas the second sentence brings noise to some extent.
In the end, proper strategies must be adopted to avoid the interference of noise information.

Indeed, there exist several attempts 
to utilize global information besides single sentence for NER~\cite{akbik2019pooled,luo2019hierarchical,2020Leveraging, Chen2020Named}.
However, these methods still easily fail to achieve
the above task due to the following facts.

First, 
they~\cite{akbik2019pooled, luo2019hierarchical, Chen2020Named} don't provide a sufficiently effective method to address the potential noise problems while introducing global information. Second, they~\cite{akbik2019pooled, luo2019hierarchical, 2020Leveraging} only utilize extra information at the word-level, but ignore the modeling at the sentence level.


In this paper, we propose \textsf{GCDoc} that exploit global contextual information for document-level NER from both word and sentence level perspectives.
At the word level,
a simple but effective document graph is constructed to model the connection between words that reoccur in document, then the contextualized representation of each word is enriched. 
To avoid the interference of noise information, we further propose two strategies.
We first apply the epistemic uncertainty theory to find out tokens whose representations are less reliable,
thereby helping prune the document graph.
Then we explore a \emph{selective auxiliary classifier} for effectively capturing the importance of different neighboring nodes.
At the sentence level, we propose a cross-sentence module to appropriately model wider context.
Three independent sentence encoders are utilized to encode the current sentence, its previous sentences as well as its
next sentences with certain window size to be discussed later.
The enhanced sentence representations are assigned to each token at the embedding layer.
To verify the effectiveness of our proposed model, we conduct extensive experiments on two benchmark NER datasets. Experimental results suggest that our model can achieve state-of-the-art performance, demonstrating that our model truly exploits useful global contextual information.

The main contributions are as follows.
\begin{itemize}
\item We propose a novel model named GCDoc to effectively exploit global contextual information by introducing document graph and cross-sentence module.
\item Extensive experiments on two benchmark NER datasets verify the effectiveness of our proposed model.
\item We conduct an elaborate analysis with various components in GCDoc and analyze the computation complexity to investigate our model in detail.
\end{itemize}

\section{Related Work}
Most of conventional high performance NER systems are based on classical statistical machine learning models,
such as Hidden Markov Model (HMM)~\cite{rabiner1989tutorial},
Conditional Random Field (CRF)~\cite{lafferty2001conditional,Mccallum2003Early,2013Chinese}, Support Vector Machine (SVM)~\cite{kudoh2000use} and \etc
Although great success has been achieved,
these methods heavily relying on handcrafted features and external domain-specific knowledge.
With the rise of deep learning,
many research efforts have been conducted on neural network based approaches
to automatically learning feature representations for NER tasks.

\paratitle{LSTM-based NER}.
Currently, LSTM is widely adpoted as the context encoder for most NER models.
Huang \etal \cite{huang2015bidirectional} initially employ a Bi-directional LSTM (Bi-LSTM) and CRF model for NER for better capturing long-range dependency features and thus achieve excellent performances.
Subsequently, the proposed architecture is widely used for various sequence labeling tasks~\cite{Ma2016End,Lample2016Neural}.
Several researches~\cite{Ma2016End,Lample2016Neural} further extend such model with an additional LSTM/CNN layer to encode character-level representations.
More recently, using contextualized representations from pretrained language models has been adopted to significantly improve performance~\cite{Peters2018Deep,devlin2018bert}.
However, most of these approaches focus on capturing the local context dependencies, which prevents the modeling of global context.

\paratitle{Exploiting Global Context}.
Several researches have been carried out to utilize global information besides single sentence for NER. 
Qian \etal~\cite{qian2018graphie} model non-local context dependencies in document level and aggregates information from other sentences via convolution.
Akbik \etal~\cite{akbik2019pooled} dynamically aggregate contextualized embeddings for each unique string in previous sentences, and use a pooling operation to generate a global word representation.
All these works ignore the potential noise problems while introducing global information. 
Luo \etal~\cite{2017An} adopt the attention mechanism to leverage document-level information and enforce label consistency across multiple instances of the same token.
Similarly,
Zhang \etal~\cite{zhang2018global} propose to retrieve supporting document-level context
and dynamically weight their contextual information by attention.
More recently, Luo \etal~\cite{luo2019hierarchical} propose a model that adopts memory networks to memorize the word representations in training datasets. 
Following their work, Gui \etal~\cite{2020Leveraging} adopt similar memory networks to record document-level information and explicitly model the document-level label consistency between the same token sequences.
Though attention operation is called in these models to compute the weight of document-level representations,
it’s not clear enough which context information should be more reliable,
thus the ability of these methods to filtering noise information is limited.
Additionally, all the above work exploit extra information at the word-level, ignoring the modeling at the sentence level.
Chen \etal~\cite{Chen2020Named} propose to capture interactions between sentences within the same document via a multi-head self attention network, while ignoring exploiting global information at the word level.
Different from their work, we propose a more comprehensive approach to exploit global context information, which has been demonstrated more effective on NER task.

\section{Preliminary}
\label{sec:prelimi}

In this section, we introduce the problem statement and
then briefly describe a widely-adopted and well-known method for NER,
\ie \emph{Bi-LSTM-CRF},
which is also a baseline in our experiment.
And then we briefly present the theories of representing model uncertainty,
which we apply for pruning the document graph module.



\subsection{Problem Formulation}
Let $\matrix{X}_{\mathcal{D}} = \{x_i\}_{m}$ be a sequence composed of $m$ 
tokens that represents a document $\mathcal{D}$ 
and $\matrix{Y}_{\mathcal{D}}=\{y_i\}_{m}$ be its corresponding sequence of tags
over $\matrix{X}_{\mathcal{D}}$.
%
Specifically,
each sentence $s_i$ in $\mathcal{D}$
is denoted by
$s_i = \{x^{i}_j\}_{n_i}$, where $n_i$ is the length of $s_i$
and $x^{i}_j$ indicates the $j$-th token of $s_i$.
%
Formally,
given a document $\mathcal{D}$,
the problem of document-level name entity recognition (NER) is to learn a parameterized ($\theta$) prediction function,
\ie
\begin{math}
  f_{\theta}\colon \matrix{X}_{\mathcal{D}} \mapsto \matrix{Y}_{\mathcal{D}}
\end{math}
from input tokens to NER labels over the entire document $\mathcal{D}$.

\subsection{Baseline:
Bi-LSTM-CRF Model}
\label{sec:prelimi-bilstm-crf}
In Bi-LSTM-CRF Model,
an input sequence
$\matrix{X}=[x_1,x_2,\ldots,x_n]$
is typically encoded into a sequence of low-dimensional distributed dense vectors,
in which each element is formulated as
a concatenation of pre-trained $\vec{w}_i$
and character-level $\vec{c}_i$ word embeddings $\vec{x}_i=[\vec{w}_i, \vec{c}_i]$.
Then, a \emph{context} encoder is employed (\ie \emph{Bi-directional LSTM})
to encode the embedded $\matrix{X}$ into a sequence of hidden states $\matrix{H}=[\vec{h}_1,\vec{h}_2,\ldots,\vec{h}_n]$
for capturing the \emph{local} context dependencies within sequence,
namely,
\begin{gather}
\label{pre_lstm_1}
\stackrel{\rightarrow}{\vec{h}_i} = \mbox{LSTM}(\stackrel{\longrightarrow}{\vec{h}_{i-1}},[\vec{w}_i;\vec{c}_i]) \\
\stackrel{\leftarrow}{\vec{h}_i} = \mbox{LSTM}(\stackrel{\longleftarrow}{\vec{h}_{i-1}},[\vec{w}_i;\vec{c}_i])\\
\vec{h}_i = [\stackrel{\rightarrow}{\vec{h}_i},\stackrel{\leftarrow}{\vec{h}_i}],
\label{pre_lstm_3}
\end{gather}
where
$\stackrel{\rightarrow}{\vec{h}_i}$ and $\stackrel{\leftarrow}{\vec{h}_i}$ are the hidden states incorporating past and future contexts of $\vec{x}_i$, respectively.

Subsequently, conditional random field (CRF~\cite{lafferty2001conditional}),
which is commonly used as a tag decoder,
is employed for jointly modeling and predicting the final tags
based on the output (\ie $\matrix{H}$) of the \emph{context} encoder,
which can be transformed to
the probability of label sequence $\vec{Y}$ being the correct tags
to the given  $\matrix{X}$, namely,

{
\begin{small}
\begin{gather}
\label{1}
  \Pr(\vec{Y}|\matrix{X}) = \frac{
  \prod_{y_j\in \vec{Y}}\phi(y_{j-1},y_j,\vec{h}_j)}
{
\sum_{y^{'}_j\in \vec{Y}^{'};\vec{Y}^{'}\in{\matrix{Y}(\matrix{X})}}
%
\prod_{j=1}^n\phi(y^{'}_{j-1},y^{'}_j,\vec{h}_j)} \\
\label{2}
	\phi(y_{j-1},y_j,\vec{h}_j)\!=\!\exp(\matrix{W}_{y_{j-1},y_{j}}\vec{h}_j + b_{y_{j-1},y_j}),
\end{gather}
\end{small}
}
where $\phi(.)$ is the score function;
$\matrix{Y}(\matrix{X})$ is the set of possible label sequences for $\matrix{X}$;
$\matrix{W}_{y_{j-1},y_{j}}$ and  $b_{y_{j-1},y_j}$
indicate the weight matrix and bias corresponding to $(y_{j-1},y_j)$.
Then, a likelihood function $\mathcal{L}_m$ is employed
to maximize the negative log probability
of ground-truth sentences for training,
\begin{equation}
\label{loss_main}
\mathcal{L}_m = -\sum_{\matrix{X}\in \mathcal{X};\vec{Y}\in \mathcal{Y}}\log \Pr(\vec{Y}|\matrix{X}),
\end{equation}
where $\mathcal{X}$ denotes the set of training instances, and $\mathcal{Y}$ indicates the corresponding
tag set.

\subsection{Representing Model Uncertainty}

Epistemic (model) uncertainty explains the uncertainty in the model parameters.
In this paper, we adopt the epistemic uncertainty to find out tokens whose labels are likely to be incorrectly predicted by the model,
thereby pruning the document graph to effectively prevent the further spread of noise information.

Bayesian probability theory provides mathematical tools to estimate model uncertainty, 
but these usually lead to high computational cost. 
Recent advances in variational inference have introduced new techniques into this field, which are used to obtain new approximations
 for Bayesian neural networks. 
Among these, Monte Carlo dropout~\cite{2015Dropout} is one of the simplest and most effective methods, which 
requires little modification to the original model and is applicable to any network architecture that makes use of dropout.

Given the dataset $\mathcal{D}$ with training inputs $\matrix{X} = \{x_1,\ldots, x_n\}$ and their corresponding labels $\matrix{Y} = \{y_1, \ldots, y_n\}$,
the objective of Bayesian inference is estimating $p(\vec{w}|\mathcal{D})$, \ie the posterior distribution of the model parameters $\vec{w}$ given the dataset $\mathcal{D}$.
Then the prediction for a new input $x^*$ is obtained as follows:
\begin{equation}
\label{eq:bayes}
p(y^*|x^*,\mathcal{D}) = \int p(y^*|x^*,\vec{w})p(\vec{w}|\mathcal{D})d\vec{w}.
\end{equation}

Since the posterior distribution $p(\vec{w}|\mathcal{D})$ is intractable, Monte Carlo dropout adopts the variational inference approach to use $q_\theta(\vec{w})$,
a distribution over matrices whose columns are randomly set to zero, to approximate the intractable posterior. 
Then the minimisation objective is the Kullback-Leibler (KL) divergence
between the approximate posterior $q_\theta(\vec{w})$ and the true posterior $p(\vec{w}|\mathcal{D})$.

With Monte Carlo sampling strategy, the integral of Eq. \ref{eq:bayes} can be approximated as follows:
\begin{equation}
\label{eq:bayes2}
p(y^*|x^*,\mathcal{D}) \approx \sum_{t=1}^{T} p(y^*|x^*,\vec{w}_t)q_\theta(\vec{w}_t),
\end{equation}
where $\vec{w}_t \sim q_\theta(\vec{w})$ and $T$ is the number of sampling.
In fact, this is equivalent to performing $T$ stochastic forward passes
through the network and averaging the results.
In contrast to standard dropout that only works during training stage, the Monte Carlo dropouts are also activated at test time.  
And the predictive uncertainty can be estimated by collecting the results of stochastic forward passes through the model and summarizing the predictive variance.

\begin{figure*}
  \centering
	\includegraphics[width=1.0 \textwidth]{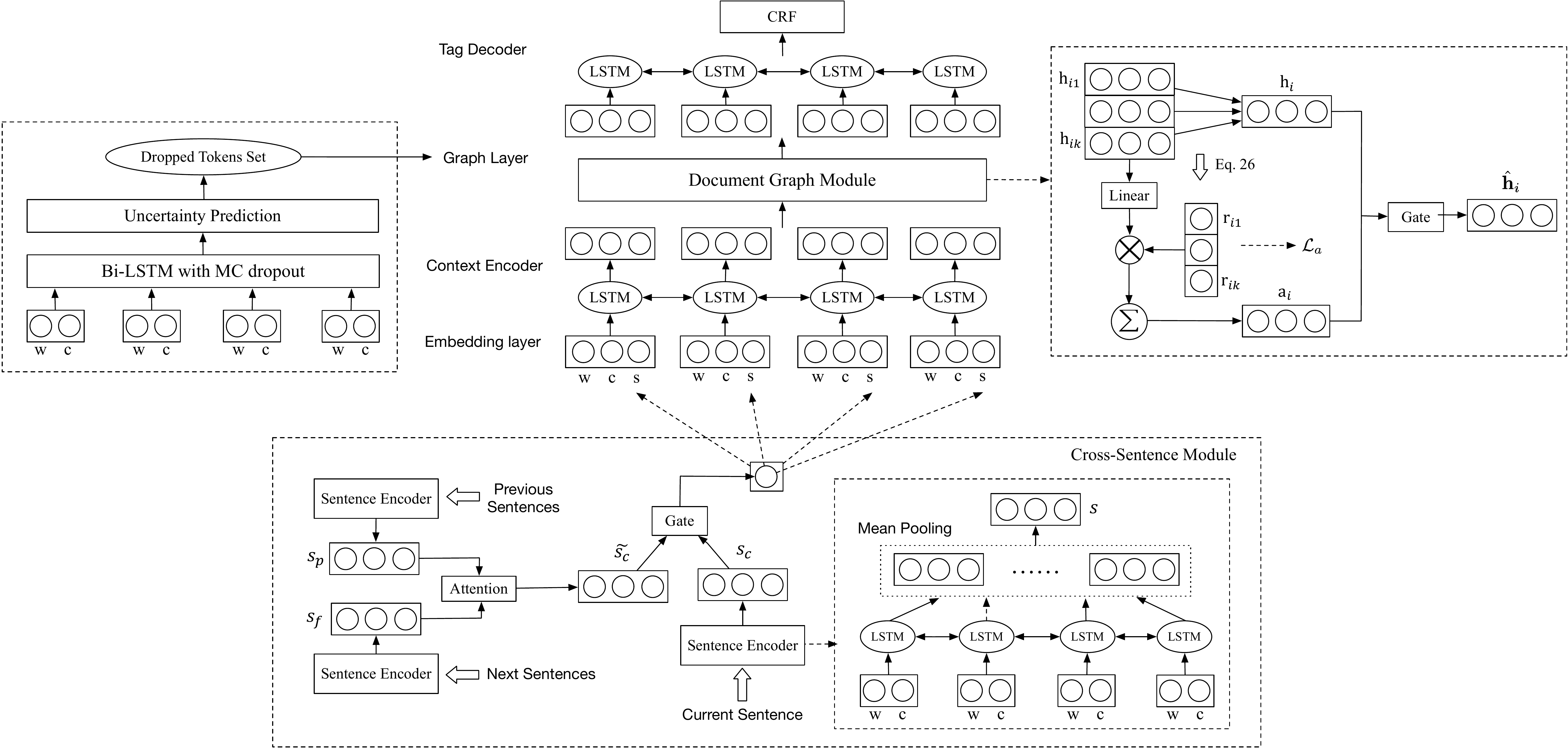}
	\caption{The architecture of our proposed NER model. The upper center side shows the overall architecture, including embedding layer, context encoder, graph layer and tag decoder. We show the uncertainty prediction sub-module on the upper left side, and on the upper right side, we show the details of the document graph module. Note that the vectors $\vec{h}_{i1} \ldots \vec{h}_{ik}$ represents the neighbor nodes' representations of node $v_i$. On the bottom we present the cross-sentence module that generates the enhanced sentence-level representation.}
	\label{fig:main}
\end{figure*}

\section{Proposed Model}

\subsection{Overview}
In this section, we focus on fully exploiting the document-level global
contexts for enhancing the word-level representation capability
of tokens,
and thus propose a novel NER model named \textbf{GCDoc},
and then we proceed to structure the following sections with more details about
each submodule of \textbf{GCDoc},
which is under the Bi-LSTM-CRF framework (\rf Section \ref{sec:prelimi-bilstm-crf})
and shown in Figure \ref{fig:main}.

\paratitle{Token Representation}.
Given a document composed of a set of tokens,
\ie $\matrix{X}_{\mathcal{D}} = [x_1,\cdots,x_m]$,
the embedding layer is formulated as a concatenation of three
different types of embeddings for obtaining
the \emph{word}-level (pre-trained $\vec{w}_i$ and character-level $\vec{c}_i$)
and the \emph{sentence}-level (\ie $\vec{s}_i$) simultaneously,
\begin{equation}
\label{eq-word-concation}
  \vec{x}_i = [\vec{w}_i;\vec{c}_i;\vec{s}_i].
\end{equation}

Generally,
the adjacent sentences
may be naturally topically relevant
to the current sentence within a same document,
and thus
we propose a novel cross-sentence contextual embedding module (bottom part in Figure \ref{fig:main}, \rf Section~\ref{method:cross-sentence})
for encoding the \emph{global}-level context information at sentence-level
to enhance the representations of tokens within the current sentence.

\paratitle{Context Encoder}.
Next, the concatenated token embeddings $\vec{x}_i=[\vec{w}_i;\vec{c}_i;\vec{s}_i]$ (instead of $[\vec{w}_i;\vec{c}_i]$) is
fed into a Bi-directional LSTM model
to capture the \emph{local} context dependencies within sequence,
and then we obtain a sequence hidden states $\matrix{H}=[\vec{h}_1,\vec{h}_2,\ldots,\vec{h}_n]$.

For enriching the contextual representation of each token,
we develop a document graph module based on gated graph neural networks
to capture the non-local dependencies
between tokens across sentences,
namely,

\begin{equation}
\label{eq-gnn}
    \hat{h}_{i}=GGNN(\vec{h}_{i}),
\end{equation}
where $GGNN(.)$ is the output of graph neural network for learning the contextualized representations of tokens.
Then, the output of Eq~\eqref{eq-gnn} can be fed into the tag decoder for training
via Eq~\eqref{1}, Eq~\eqref{2} and Eq~\eqref{loss_main}.


\subsection{Cross-Sentence Contextual Embedding Module}
\label{method:cross-sentence}
Generally, the adjacent sentences may be topically relevant to
the current sentence within a same document naturally, and thus we propose a novel cross-sentence contextual embedding module.
To this end, in this section we propose a cross-sentence module,
which treats the adjacent sentences as auxiliary context at sentence-level to
enhance the representation capability of the given sentence.
Similar to
\cite{mikolov2013efficient},
we only take account of the cross-sentence context information
within a certain range (\eg $k$) before and after the current sentence,
namely,
we concatenate the $k$ sentences before/after the current sentence
for capturing cross-sentence context information.

Without loss of generality,
suppose the representations of the current
sentence, the past context representation
of sentences and
the future context representation of sentences
are denoted by
$\vec{s}_c$,
$\vec{s}_p$
and
$\vec{s}_f$, respectively.
Furthermore,
we also introduce an attention mechanism for
measuring the relevance of such two auxiliary representations to the current one,
and thus a weighted sum ($\tilde{\vec{s}}_c$) of such two representations
is calculated for the representation of the current sentence,
namely,
\begin{equation}
\label{method_cross_sentence_begin}
\tilde{\vec{s}}_c={a_p{\vec{s}_p}} + {a_f{\vec{s}_f}},
\end{equation}
where $a_p$ and $a_f$ are
the learned attentions of the past and the future context information
to the current sentence,
which are used to control the contributions of such two context information
and computed based on \emph{softmax},
\begin{gather}
  a_p = \mbox{softmax}(f(\vec{s}_c,\vec{s}_p))\\
  a_f = \mbox{softmax}(f(\vec{s}_c,\vec{s}_f)),
\end{gather}
where $f(\vec{s}_i,\vec{s}_j)$ is
a compatibility function to calculate the \emph{pair-wise} similarity between two sentences,
which is computed by
\begin{equation}
f(\vec{s}_i,\vec{s}_j)=\vec{w}^{\top}\sigma\left(\matrix{W_s^{(i)}}\vec{s}_{i}+\matrix{W_s^{(j)}}\vec{s}_{j}+\vec{b}\right),
\end{equation}
where $\sigma(\cdot)$ is an activation function;
$\matrix{W_s^{(i)}}$, $\matrix{W_s^{(j)}} \in \mathbb{R}^{d_s\times{d_s}}$ are the learnable weight matrices;
$\vec{w}^{\top} \in \mathbb{R}^{d_s}$ is a weight vector, and $\vec{b}$ denotes
the bias vector.

Finally, a gating mechanism
is employed to
decide the amount of cross-sentence context information ($\tilde{\vec{s}}_c$) that should be incorporated and fused with the initial sentence representation $\vec{s}_c$:
\begin{equation}
    \vec{s'}_c= \lambda\odot\vec{s}_c +(1-\lambda)\odot\tilde{\vec{s}}_c,
\end{equation}
where $\odot$ represents element\-wise multiplication,
and $\lambda$ is the trade-off parameter,
which is calculated by
\begin{equation}
\label{method_cross_sentence_end}
   \lambda = \mbox{sigmoid}\left(\matrix{W_g^{(3)}}\tanh(\matrix{W_g^{(1)}}\tilde{\vec{s}}_c+\matrix{W_g^{(2)}}\vec{s}_c)\right),
\end{equation}
where $\matrix{W_g^{(1)}}, \matrix{W_g^{(2)}}, \matrix{W_g^{(3)}} \in \mathbb{R}^{d_s\times{d_s}}$ are trainable weight matrices.

\paratitle{Remark}.
The enhanced sentence-level representation $\vec{s'}_c \in \mathbb{R}^{d_s}$ is then concatenated with
the word-level embeddings of tokens (\rf Eq~\eqref{eq-word-concation}) and fed into the context encoder.
In particular,
%
For each sentence,
we adopt Bi-LSTM
as the basic architecture for sentence embedding.
Specifically, for each sentence $s$,
%
a sequence of hidden states $\vec{H}_{s}=\{\vec{h}^{s}_{i}\} \in \mathbb{R}^{d_s}$
are generated by Bi-LSTM model according to Eq~\eqref{pre_lstm_1}-\eqref{pre_lstm_3},
%
and then the sentence representation
$\vec{s}$ is computed by mean pooling over all hidden states $\vec{H}_{s}$,
that is,
\begin{equation}
\label{method_sentence_encoder}
\vec{s} = \frac{1}{n}\sum_{i=1}^{n}\vec{h}^{s}_{i},
\end{equation}
where $n$ is the length of the sequence.

\begin{figure}
\hspace{-0.4cm}
  \centering
	\includegraphics[width=0.5\textwidth]{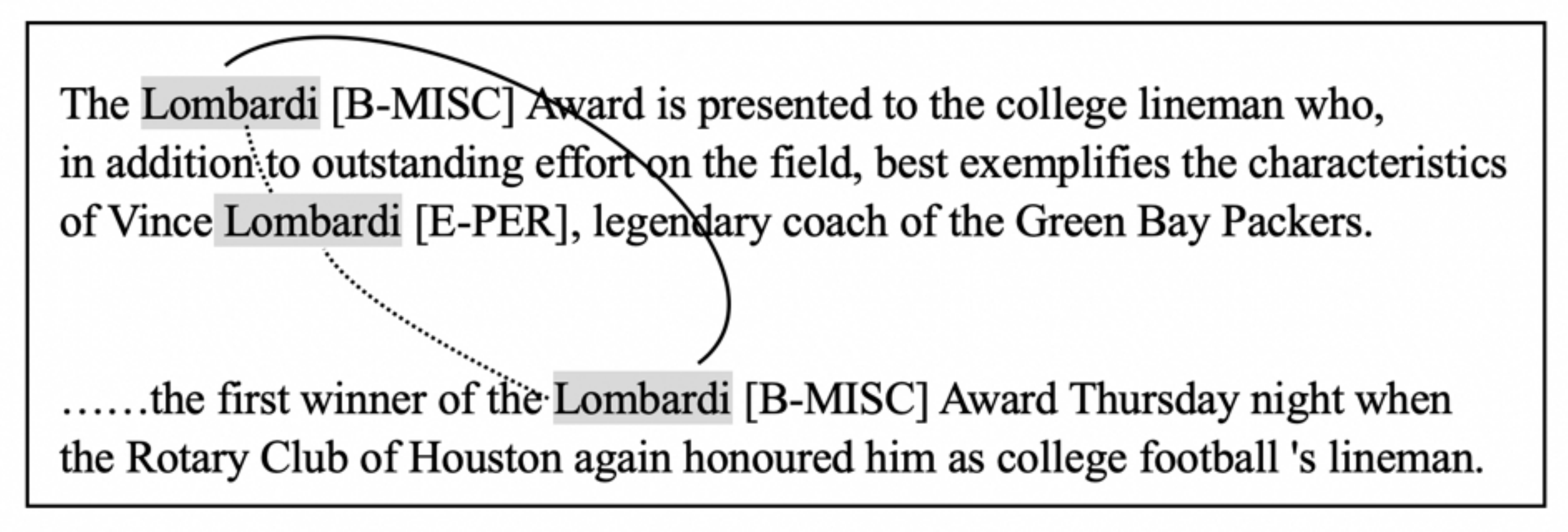}
 \caption{Illustration for constructing document graph. Words appearing repeatedly in a document are connected, and for clarity, we only adopt the word ``Lombardi" as an example. 
 Aside from the edge (solid line) connecting two nodes belonging to the same entity type, there are edges (dashed lines) that may introduce noise.
 }
	\label{fig:graph}
\end{figure}

\subsection{Gated Graph Neural Networks based Context Encoder}
As mentioned,
an intuition is that the entity mentions appearing repeatedly in a document
are more likely to be of same entity category.
To this end,
in this section,
we develop a simple but effective
graph layer based on a word-level graph,
which is built over
the re-occurrent tokens to capture the non-local dependencies
among them.

\paratitle{Graph Construction}.
Given a document $\matrix{X}_{\mathcal{D}}=\{x_i\}_{m}$,
a recurrent graph is defined as an undirected graph
$\mathcal{G}_\mathcal{D} = (\mathcal{V}_\mathcal{D}, \mathcal{E}_\mathcal{D})$,
where each node $v_i\in \mathcal{V}_\mathcal{D}$ denotes
a recurrent token,
each edge $e_{ij}=(v_i,v_j)\in \mathcal{E}_\mathcal{D}$
is a connection of every two same case-insensitive tokens (\rf Figure~\ref{fig:graph}),
which is used to model the re-occurrent context information
of tokens across sentences, as the local context of a token in  a sentence may be ambiguous or limited.

\paratitle{Node Representation Learning}.
To learn the node representations of tokens,
we adopt graph neural networks
to generate the local permutation-invariant aggregation on the neighborhood of a node in a graph,
%
and thus the non-local context information of tokens can be efficiently captured.

Specifically, for each node $v_{i}$,
the information propagation process between different nodes can be formalized as:
\begin{equation}
\label{aggre_neighbor}
\vec{a}_i =   \mbox{ReLU}\left( \frac{1}{|\mathcal{N}(i)|}\sum_{j \in \mathcal{N}(i)}\matrix{W}_a\vec{h}_j+\vec{b}_a \right),
\end{equation}
where $\mathcal{N}(i)$ denotes the set of neighbors of $v_i$, which does not include $v_i$ itself;
$\matrix{W}_a \in \mathbb{R}^{d \times d}$ indict the weight matrices and $\vec{b} \in \mathbb{R}^d$ denotes the bias vector;
$\vec{h}_j$ is the hidden state of the neighbor node $v_j\in \mathcal{N}(i)$.

Through Eq~\eqref{aggre_neighbor},
we can easily
aggregate the neighbours' information
to enrich the context representation of node $v_i$
via the constrained edges ($e_{ij}\in\mathcal{E}_\mathcal{D}$).
Inspired by the gated graph neural networks,
the enhanced representation $\vec{a}_i$ is 
combined with its initial embedding ($\vec{h}_i$) via a gating mechanism,
which is employed for deciding the amount of aggregated context information to be incorporated,
namely,
\begin{gather}
\label{eq:ggnn_begin}
\hat{h}_i = (1-\vec{z}_i)\odot\vec{h}_i+\vec{z}_i\odot\tilde{\vec{h}_i}\\
\tilde{\vec{h}}_i = \tanh(\matrix{W}_o\vec{a}_i+\matrix{U}_o(\vec{r}_i\odot \vec{h}_i))\\
\vec{z}_i = \sigma(\matrix{W}_z\vec{a}_i+\matrix{U}_z\vec{h}_i)\\
\label{eq:ggnn_end}
\vec{r}_i = \sigma(\matrix{W}_r\vec{a}_i+\matrix{U}_r\vec{h}_i),
\end{gather}
where $\matrix{W}_z, \matrix{W}_r, \matrix{W}_o, \matrix{U}_z, \matrix{U}_r, \matrix{U}_o\in \mathbb{R}^{d \times d}$ are learnable weight matrices, $\sigma(\cdot)$ denotes the logistic sigmoid function and $\odot$ represents element\-wise multiplication. $\vec{z}_i$ and $\vec{r}_i$ are update gate and reset gate, respectively.
The output of Eq \eqref{eq:ggnn_begin} $\hat{h}_i$ is the final output of our document graph module (\rf Eq \eqref{eq-gnn}), which will be fed into the tag decoder for generating the tag sequence.

Furthermore,
directly aggregating the contextual information may contain noise,
since not all re-occurrent tokens can provide reliable context information for disambiguating others.
As such, we propose two strategies to avoid introducing noise information in the document graph module.
First we apply the epistemic uncertainty to find out tokens whose labels are likely to be incorrectly predicted,
thereby pruning the document graph.
Then we adopt a selective auxiliary classifier for distinguishing the entity categories of two different nodes, thus 
guiding the process of calculating edge weights.

\paratitle{(1) Document Graph Pruning}.
We suppose that tokens whose labels are more likely to be correctly predicted by the model will have more reliable contextual representations, and will be more helpful in the graph module for disambiguating other tokens.
In this paper, we use Monte Carlo dropout to estimate the model uncertainty for indicating whether model predictions 
are likely to be correct.
Specifically, we obtain the uncertainty value of each token through an independent sub-module. Then we set a threshold $\mathcal{T}$, and put tokens whose uncertainty value is greater than the threshold into a set $Drop$.
Since the context information of these tokens should be less reliable,
we prune the graph module by ignoring the impact of these tokens on their neighboring nodes.
Thus Eq \ref{aggre_neighbor} is rewritten as follows,
\begin{equation}
\label{aggre_neighbor_2}
\vec{a}_i = \mbox{ReLU}\left( \frac{1}{Num(i)}\sum_{j \in \mathcal{N}(i) \& j \notin Drop}\matrix{W}_a\vec{h}_j+\vec{b}_a \right),
\end{equation}
where $Num(i)$ denotes the the number of neighbor nodes of node $i$ that actually participating in the operation.
                
As for the model uncertainty prediction, we apply the Monte Carlo dropout and adopt a independent sub-module that
predicts labels of each token by a simple architecture composed of a Bi-LSTM and dense layer.
The forward pass of the Bi-LSTM is run $T$ times with the same inputs by using the 
approximated posterior $q_\theta(\vec{w})$ in Eq \ref{eq:bayes2}.
Then given the output representation $\vec{h}_i$ of the Bi-LSTM layer, 
the probability distribution of the $i$-th word’s label can be obtained by a fully connected layer and a final softmax function,
\begin{equation}
\vec{p}_i \approx \sum_{t=1}^{T} \mbox{softmax}(\matrix{W}^\top\vec{h}_i | \vec{w}_t),
\end{equation}
where $\vec{w}_t \sim q_\theta(\vec{w})$, $\matrix{W} \in \mathbb{R}^{d \times \mathcal{C}}$, and $\mathcal{C}$ is the number of all possible labels.
Then, the uncertainty value of the token can be represented by the uncertainty of its corresponding probability vector $\vec{p}_i$,
which can be summarized using the entropy of the probability vector:
\begin{equation}
u_i = -\sum_{c=1}^{\mathcal{C}}p_c\log p_c.
\end{equation}

\paratitle{(2) Edge Weights Calculation}.
Since not all of repeated entity mentions in document belong to the same categories, which is
shown in Figure~\ref{fig:graph}, we propose a selective auxiliary classifier
for distinguishing the entity categories of two different nodes, 
which is defined as a binary classifier that takes the representation of two nodes as input and outputs whether it should be the same entity category,
which is denoted by the relation score $r_{ij}$ for any edge $e_{ij}$
and is computed by
\begin{equation}
\label{edge_weight}
r_{ij} = \sigma (\matrix{W}_c[\vec{h}_i;\vec{h}_j]+b_c),
\end{equation}
where $\matrix{W}_c\in \mathbb{R}^{2d}$ is a learnable weight matrix, $b_c$ represents the bias and $\sigma(\cdot)$ denotes the logistic sigmoid function.

In specific,
the loss function of the selective auxiliary classifier is defined as:
\begin{equation}
\mathcal{L}_a = -\sum_{i=1}^m \sum_{j \in \mathcal{N}(i)}[y_{ij}\log(r_{ij})+(1-y_{ij})\log(1-r_{ij})],
\label{equ:loss_aui}
\end{equation}
where
$y_{ij}$ is the ground truth label whose value is assigned $1$ when nodes $v_i$ and $v_j$ belong to the same entity category, and $0$ for otherwise.
As a result, Eq~\eqref{aggre_neighbor_2}
can be transformed as follows via the computed relation scores,
namely,
\begin{equation}
\label{final_aggre}
\vec{a}_i = \mbox{ReLU}\left( \frac{1}{Num(i)}\sum_{j \in \mathcal{N}(i) \& j \notin Drop}r_{ij}\left(\matrix{W}_a\vec{h}_j+\vec{b}_a\right) \right).
\end{equation}
After obtaining the enhanced representation $\vec{a}_i$ that encodes the neighbours’ information, Eq \eqref{eq:ggnn_begin}-\eqref{eq:ggnn_end} are applied to generate the final output of our document graph module $\hat{h}_i$.

The employed auxiliary classifier can be regarded as a special regularization term, which explicitly injects the supervision information into the process of calculating edge weights and helps the model select useful information in the process of aggregating neighbors.

The final loss of our model is the sum of loss $\mathcal{L}_m$ (Eq \ref{loss_main}) and loss $\mathcal{L}_a$ (Eq \ref{equ:loss_aui}),
\begin{equation}
 Loss = \mathcal{L}_m + \theta\mathcal{L}_a,
 \label{eq:final_loss}
\end{equation}
where $\theta$ is a hyper-parameter. 

\section{Experiments}

\subsection{Data Sets}

We use two benchmark NER datasets for evaluation, \ie \emph{CoNLL 2003 dataset} (CoNLL03) and \emph{OntoNotes 5.0 English datasets} (OntoNotes 5.0).
The details about corporas are shown in Table \ref{table1}.
\begin{itemize}
\item \textbf{CoNLL03}~\cite{2003Introduction} is a collection of news wire articles from the \emph{Reuters} corpus,
            which includes four different types of named entities: persons (PER), locations (LOC), organizations (ORG), and miscellaneous (MISC).
            We use the standard dataset split~\cite{Collobert2011Natural} and follow \emph{BIOES} tagging scheme (B, I, E represent the beginning, middle and end of an entity, S means a single word entity and O means non-entity).
\item \textbf{OntoNotes 5.0} is much larger than \emph{CoNLL 2003 dataset}, and consists of text from a wide variety of sources (broadcast conversation, newswire, magazine and Web text, \etc.) It is tagged with eighteen entity types, such as persons (PERSON), organizations (ORG), geopolitical entity (GPE), and law (LAW), \etc.
Following previous works~\cite{chiu2016named}, we adopt the portion of the dataset with gold-standard named entity annotations, in which the New Testaments portion is excluded.
\end{itemize}
\subsection{Network Training}
In this section, we show the details of network training. Related hyper-parameter settings are presented including initialization, optimization, and network structure. We also report the training time of our model on the two datasets.

\paratitle{Initialization.}
We initialize word embedding with $300$-dimensional GloVe~\cite{pennington2014glove} and randomly initialize $30$-dimensional character embedding.
We adopt fine-tuning strategy and modify initial word embedding during training.
All weight matrices in our model are initialized by Glorot Initialization~\cite{glorot2010understanding}, and the bias parameters are initialized with $0$.

\paratitle{Optimization.}
We train the model parameters by the \emph{mini-batch} stochastic gradient descent (SGD) with batch size $20$ and learning rate $0.01$.
The $L_2$ regularization parameter is $1e$-$8$. 
We adopt the dropout strategy to overcome the over-fitting on the input and the output
of Bi-LSTM with a fixed ratio of $0.5$.
We also use a gradient clipping~\cite{Pascanu2012On} to avoid gradient explosion problem. The threshold of the clip norm is set to $5$.
Early stopping~\cite{Lee2000Overfitting} is applied for training models according to their performances on development sets.

\paratitle{Network Structure.}
We use Bi-LSTM to learn character-level representation of words,
which will form the distributed representation for input
together with the pre-trained word embedding and sentence-level representation.
The size of hidden state of character and word-level Bi-LSTM are set to $50$ and $200$, respectively. And we fix the depth of these layers as 1 in our neural architecture. The hidden dim of sentence-level representation is set to $300$ and the window size $k$ in cross-sentence module is $2$. The threshold $\mathcal{T}$ in uncertainty prediction sub-module is set to $0.5$. The hyper-parameter $\theta$ in Eq \ref{eq:final_loss} is set to $0.1$.
In the auxiliary experiments, the output hidden states of BERT~\cite{devlin2018bert} are taken as additional contextualized embeddings, which are combined with other three types of word representations (\rf Eq \ref{eq-word-concation}).
We still adopts Bi-LSTM to encode the context of words instead of using BERT as encoder and fine-tune it.

\paratitle{Training Time.}
We implement our model based on the PyTorch library and the training process has been conducted on one GeForce GTX 1080 Ti GPU.
The model training completes in about 1.6 hours on the CoNLL03 dataset and about 6.7 hours on the OntoNotes 5.0 dataset.

\setlength{\tabcolsep}{1pt}
\begin{table}[!t]
	\footnotesize
	\begin{center}
		\begin{tabular}{c|c|c|c|c}
			\hline
			\bf Corpus&\bf Type&\bf Train&\bf Dev&\bf Test\\
			\hline
			\hline
			{\multirow{2}{*}{\emph{CoNLL03}}}&Sentences&14,987&3,466&3,684\\
			&Tokens&204,567&51,578&46,666\\
			
			\hline
			{\multirow{2}{*}{\emph{OntoNotes 5.0}}}&Sentences&59,924&8,528&8,262\\
			&Tokens&1,088,503&147,724&152,728\\
			\hline
		\end{tabular}
	\end{center}
	\caption{Statistics of \emph{CoNLL03} and \emph{OntoNotes 5.0}.}
    \label{table1}
\end{table}

\subsection{Evaluation Results and Analysis}
\begin{table}[!t]
	\footnotesize
		\begin{tabular}{c|c|c|c}
			\hline
			{\multirow{2}{*}{\bf Category}} &
			{\multirow{2}{*}{\bf Index \& Model}}&\multicolumn{2}{c}{\bf $F_1$-score} \\
			\cline{3-4}~&~&\bf Type&\bf Value($\pm $$\mbox{std}^{1}$)\\
			\hline
			{\multirow{6}{*}{Sentence-level}} & \quad\quad Lample \etal, 2016 \cite{Lample2016Neural}\quad\quad&reported&90.94\\
			\cline{2-4}
			~ &\quad\quad Ma and Hovy, 2016 \cite{Ma2016End}\quad\quad&reported&91.21\\
			\cline{2-4}
			~ &\quad\quad Yang \etal, 2017 \cite{Yang2016Transfer}\quad\quad&reported&91.26\\
			\cline{2-4}
			~&{\multirow{2}{*}{{\quad\quad Liu \etal, 2018 \cite{Liu2017Empower}}\quad\quad}}&avg&91.24$\pm$0.12\\
			\cline{3-4}
			~&~&max&91.35\\
			\cline{2-4}
			~&\quad\quad Ye and Ling, 2018 \cite{ye2018hybrid}\quad\quad&reported&91.38\\
			\hline
			{\multirow{5}{*}{Document-level}} &
			{\multirow{2}{*}{{\quad\quad Zhang \etal, 2018 \cite{zhang2018global}}\quad\quad}}&avg&91.64\\
			\cline{3-4}
			~&~&max&91.81\\
			\cline{2-4}
			~& \quad\quad Qian \etal, 2019 \cite{qian2018graphie}\quad\quad&reported&91.74\\
			\cline{2-4}
			~& \quad\quad Luo \etal, 2020 \cite{luo2019hierarchical}\quad\quad&reported&91.96$\pm$0.03\\
			\cline{2-4}
			~& \quad\quad Gui \etal, 2020 \cite{2020Leveraging}\quad\quad&reported&92.13\\
			\hline
			~& {\multirow{2}{*}{$\mbox{Bi-LSTM-CRF}^{2}$~\cite{Lample2016Neural}}}&avg&91.01$\pm$0.21\\
			\cline{3-4}~&~&max&91.27\\
			\cline{2-4}
			~&{\multirow{2}{*}{GCDoc$\ast$}}&avg&92.22$\pm$0.02\\
			\cline{3-4}~&~&max&\textbf{92.26}\\
			\hline
			\hline
			\multicolumn{4}{c}{\bf + Language Models / External knowledge}\\
			\hline
			{\multirow{4}{*}{Sentence-level}} &		
			\quad\quad Peters \etal, 2017 \cite{Peters2017Semi}&reported&91.93$\pm$0.19\\
			\cline{2-4}
			~& \quad\quad Peters \etal, 2018 \cite{Peters2018Deep}&reported&92.22$\pm$0.10\\
			\cline{2-4}
			~& \quad\quad Akbik \etal, 2018 \cite{akbik2018contextual}&reported&92.61$\pm$0.09\\
			\cline{2-4}
			~& \quad\quad Devlin \etal, 2018 \cite{devlin2018bert}&reported&92.80\\
			\hline
			{\multirow{4}{*}{Document-level}} &	
			\quad\quad Akbik \etal, 2019 \cite{akbik2019pooled}&reported&93.18$\pm$0.09\\
			\cline{2-4}
			~&\quad\quad Chen \etal, 2020 \cite{Chen2020Named}\quad\quad&reported&92.68\\
			\cline{2-4}
			~&\quad\quad Luo \etal, 2020 \cite{luo2019hierarchical}\quad\quad&reported&93.37$\pm$0.04\\
			\cline{2-4}
			~&\quad\quad Gui \etal, 2020 \cite{2020Leveraging}\quad\quad&reported&93.05\\
			\hline
			~&{\multirow{2}{*}{GCDoc+$BERT_{LARGE}$}}&avg&93.40$\pm$0.02\\
						\cline{3-4}~&~&max&\textbf{93.42}\\
			\hline	
		\end{tabular}
    \begin{tablenotes}
        \scriptsize
        \item[]{$^{1}~std$ means Standard Deviation.}
        \item[]{$^{2}$~Here we re-implement the classical Bi-LSTM-CRF model using the same model setting and optimization method with our model.}
    \end{tablenotes}
    \caption{Comparison of overall performance on CoNLL 2003 NER task.
$\ast$ indicates statistical significance on the test dataset against Bi-LSTM-CRF by
 a paired t-test with $p<0.01$. }
    \label{conll03}
\end{table}

\subsubsection{Over Performance}
This experiment is to evaluate the effectiveness of our approach on different datasets.
Specifically, we report standard $F_1$-score for evaluation.
In order to enhance the fairness of the comparisons and verify the solidity of our improvement, we rerun $3$ times with different random initialization and report both average and max results of our proposed model.
As for the previous methods, we will show their reported results.
Note that these models use the same methods of dataset split and evaluation metrics calculation as ours, which can ensure the accuracy and fairness of the comparison.
The results are given in Table \ref{conll03} and Table \ref{table_ontonotes}, respectively.

On CoNLL03 dataset, we compare our model with the state-of-the-art models which can be categorized into sentence-level models and document-level models.
The listed sentence-level models are usually popular baselines for most subsequent work in this field~\cite{Lample2016Neural,Ma2016End,Yang2016Transfer,Liu2017Empower,ye2018hybrid}.
And the document-level models are recent work that also utilize global information besides single sentence for NER as well~\cite{zhang2018global,qian2018graphie,luo2019hierarchical,2020Leveraging,akbik2019pooled,Chen2020Named}.
Besides, we leverage the pre-trained language model BERT (large version) as an external resource for fair comparison with other models that also use external knowledge~\cite{Peters2017Semi,Peters2018Deep,akbik2018contextual}
On CoNLL03 dataset, our model achieves $92.22\%$ $F_1$-score without external knowledge and $93.40\%$ with BERT. Even compared with the more recent top-performance baseline~\cite{2020Leveraging}, our model achieves a better result with an improvement of $0.09\%$ and $0.35\%$.
Considering that the CoNLL03 dataset is relatively small, we further conduct experiment on a much more large OntoNotes 5.0 dataset.
We compare our model with previous methods that also reported results on it~\cite{chen2019grn,luo2019hierarchical,akbik2019pooled}.
As shown in Table \ref{table_ontonotes}, our model shows a significant advantage on this dataset, which achieves $88.32\% F_1$ score without external knowledge and $90.49\%$ with BERT, consistently outperforming all previous baseline results substantially.
Besides, the std (Standard Deviation) value of our model is smaller than the one of Bi-LSTM-CRF, which demonstrates our proposed method is more stable and robust.
In Tabel \ref{table_prf}, we show the comparison results of the detailed metrics (including precision, recall and $F_1$) of GCDoc and Bi-LSTM-CRF on the two datasets.
We can indicate from the table that compared with Bi-LSTM-CRF, the evaluation results of GCDoc under all metrics are significantly improved. 
And the score of recall is improved even more compared with precision.

Overall, the comparisons on these two benchmark datasets well demonstrate that our model can truly leverage the document-level contextual information to enhance NER tasks without the support from external knowledge.

\begin{table}[!t]
	\footnotesize
	\begin{center}
		\begin{tabular}{c|c|c|c}
			\hline
			{\multirow{2}{*}{\bf Category}} &
			{\multirow{2}{*}{\bf Index \& Model}}&\multicolumn{2}{c}{\bf $F_1$-score} \\
			\cline{3-4}~&~&\bf Type&\bf Value($\pm$ std)\\
			\hline
			{\multirow{4}{*}{Sentence-level}} &
			\quad\quad Chiu and Nichols, 2016 \cite{chiu2016named}\quad\quad &reported&86.28$\pm$0.26\\
			\cline{2-4}
			~& \quad\quad Strubell \etal, 2017 \cite{strubell2017fast}\quad\quad &reported&86.84$\pm$0.19\\
            \cline{2-4}
			~&
            \quad\quad Li \etal, 2017 \cite{li2017leveraging}\quad\quad &reported&87.21\\	
			\cline{2-4}
			~&
			\quad\quad Chen \etal, 2019 \cite{chen2019grn}\quad\quad &reported&87.67$\pm$0.17\\	
			\hline
			{\multirow{2}{*}{Document-level}}&
			\quad\quad Qian \etal, 2019 \cite{qian2018graphie}\quad\quad&reported&87.43\\
			\cline{2-4}
			~&
			\quad\quad Luo \etal, 2020 \cite{luo2019hierarchical}\quad\quad&reported&87.98$\pm$0.05\\
			\hline
			~&{\multirow{2}{*}{Bi-LSTM-CRF~\cite{Lample2016Neural}}}&avg&87.64$\pm$0.23\\
			\cline{3-4}~&~&max&87.80\\
			\cline{2-4}
			~&
			{\multirow{2}{*}{GCDoc$\ast$}}&avg&88.32$\pm$0.04\\
			\cline{3-4}~&~&max&\textbf{88.35}\\
			\hline
			\hline
			\multicolumn{4}{c}{\bf + Language Models / External knowledge}\\
			\hline
			{\multirow{3}{*}{Sentence-level}} &
			\quad\quad Clark \etal, 2018 \cite{clark2018semi}\quad\quad &reported&88.81$\pm$0.09\\	
			\cline{2-4}
			~&\quad\quad Liu \etal, 2019 \cite{liu2019towards}&reported&89.94$\pm$0.16\\
			\cline{2-4}
			~&\quad\quad Jie and Lu, 2019 \cite{jie2019dependency}&reported&89.88\\
			\hline
			Document-level &
			\quad\quad Luo \etal, 2020 \cite{luo2019hierarchical}\quad\quad&reported&90.30\\
			\hline
			~&
			{\multirow{2}{*}{GCDoc+$BERT_{LARGE}$}}&avg&90.49$\pm$0.09\\
			\cline{3-4}~&~&max&\textbf{90.56}\\
			\hline
		\end{tabular}
	\end{center}
    \caption{Comparison of overall performance on OntoNotes 5.0 English datasets.  
$\ast$ indicates statistical significance on the test dataset against Bi-LSTM-CRF by
 a paired t-test with $p<0.01$. }
    \label{table_ontonotes}
\end{table}

\begin{table}[!t]
	\footnotesize
	\begin{center}
		\begin{tabular}{m{.11\textwidth}m{.065\textwidth}m{.05\textwidth}m{.07\textwidth}
		m{.065\textwidth}m{.05\textwidth}m{.06\textwidth}}
			\toprule
			{\multirow{2}{*}{\bf Model}} & \multicolumn{3}{c}{\bf CoNLL03} & \multicolumn{3}{c}{\bf OntoNotes 5.0} \\
			~&\bf  Precision  &\bf Recall
			&\bf  $F_1$-score 
			&\bf Precision  &\bf  Recall
			&\bf $F_1$-score\\
			\midrule
			Bi-LSTM-CRF & 90.96 & 91.07 & 91.01 & 87.58 & 87.70 & 87.64 \\   
			GCDoc & 92.12 & 92.33 & 92.22 & 87.90 & 88.74 & 88.32 \\
			$\Delta$& +1.16 & +1.26 & +1.21 & +0.32 & +1.04 & +0.68 \\
			\bottomrule
		
		\end{tabular}
	\end{center}
    \caption{Comparison of precision, recall and $F_1$-score between GCDoc and Bi-LSTM-CRF on CoNLL03 and OntoNotes 5.0 datasets.}
    \label{table_prf}
\end{table}

\subsubsection{Ablation Study}
In this section, we run experiments to dissect the relative impact of each modeling decision by ablation studies.

For better understanding the effectiveness of two key components
in our model, \ie document graph module and cross-sentence module, we conduct ablation tests where one of the two components is individually adopted. The results are reported in Table \ref{tabel_ablation1}.
The experimental results show that by adding any of the two modules, the model's performance on both datasets is significantly improved.
We discover that just adopting the document graph module can get a large gain of $0.85$\%/$0.56$\% over baseline. And by combining with the cross-sentence module, we achieve the final state-of-the-arts results, indicating the effectiveness of our proposed approach to exploit document-level contexual information at both the word and sentence level.

\begin{table}[t!]
	\footnotesize
		\begin{tabular}{p{.02\textwidth}p{.23\textwidth}m{.1\textwidth}m{.13\textwidth}}
			\toprule
			\bf No & \bf Model & \bf CoNLL03 &\bf OntoNotes 5.0 \\
            \midrule
			1&base model$\ast$&91.01$\pm$0.21&87.64$\pm$0.23\\

			2&+ Cross-Sentence Module$\ast$&91.53$\pm$0.09&87.91$\pm$0.21\\

            3&+ Document Graph Module$\ast$&91.86$\pm$0.02&88.20$\pm$0.11\\

            4&+  ALL &92.22$\pm$0.02&88.32$\pm$0.04\\
			\bottomrule
		\end{tabular}
	\caption{Ablation study of overall architechture on the two benchmark datasets.
$\ast$ indicates statistical significance on the test dataset against GCDoc by
 a paired t-test with $p<0.01$. Note that in this table, $\ast$ measures the drop in performance.}
    \label{tabel_ablation1}
\end{table}

In order to analyze the working mechanism of the document graph module in our porposed model, we conduct additional experiments on CoNLL03 dataset.
We introduce 3 baselines as shown in Table \ref{tabel_ablation2}:
(1) base model only with cross-sentence module;
(2) add document graph module but without the two strategies for avoiding noise, which means that Eq \ref{aggre_neighbor} is adopted to aggregate neighbor information, \ie takes the average of all neighbors' representations;
(3) adopt document graph pruning strategy with uncertainty prediction sub-module, \ie using Eq \ref{aggre_neighbor_2} to aggregate neighbor information;
The last one is GCDoc, which applys a selective auxiliary classifier to guide the edge weights calculation progress and further helps the model select useful information.
By comparing Model 1 and Model 2, we can find that just adding a simple document graph can get a significant improvement of $0.47\%$, which shows that modeling the connection between words that reoccur in document indeed brings much useful global contextual information.
The experiment that comparing Model 2 and Model 3 indicates that adding document graph pruning strategy  will bring further improvement, since noise information is prevented from introducing by ignoring the impact of less reliable tokens on their neighboring nodes.
And the result of comparing Model 3 and Model 4 demonstrates that the designed selective auxiliary classifier provides additional improvements, which proves its importance for capturing the weights of different neighbor nodes and filtering noise information.

In order to appropriately model a wider context beyond single sentence, we encode both the previous and next sentences to enhance the semantic information of the current sentence.
In this experiment, one of the two parts is removed from model each time and the results are shown in Table \ref{tabel_ablation3}.
In the first baseline we only encode the current sentence and utilize it as the sentence-level representation.
We can find that including either previous or next sentences contributes to an improvement over the base model, 
which verifies the effectiveness of modeling wider context and encoding neighbour sentences to improve NER performance.

\begin{table}[t!]
	\small
	\begin{center}
		\begin{tabular}{p{.06\textwidth}p{.23\textwidth}m{.13\textwidth}}
			\toprule
			\bf No & \bf Strategies & \bf $F_1$-score$\pm$std  \\
            \midrule
            1&Base&91.53$\pm$0.09\\
			2&+Doc Graph&92.00$\pm$0.07\\
            3&+Pruning&92.10$\pm$0.03\\
            4&+Edge Weights&92.22$\pm$0.02\\
			\bottomrule
		\end{tabular}
	\end{center}
	\caption{Comparison of different strategies of Document Graph Module on CoNLL03 dataset.}
    \label{tabel_ablation2}
\end{table}

\begin{figure}[ht]
	\centering
	\includegraphics[width=0.48\textwidth]{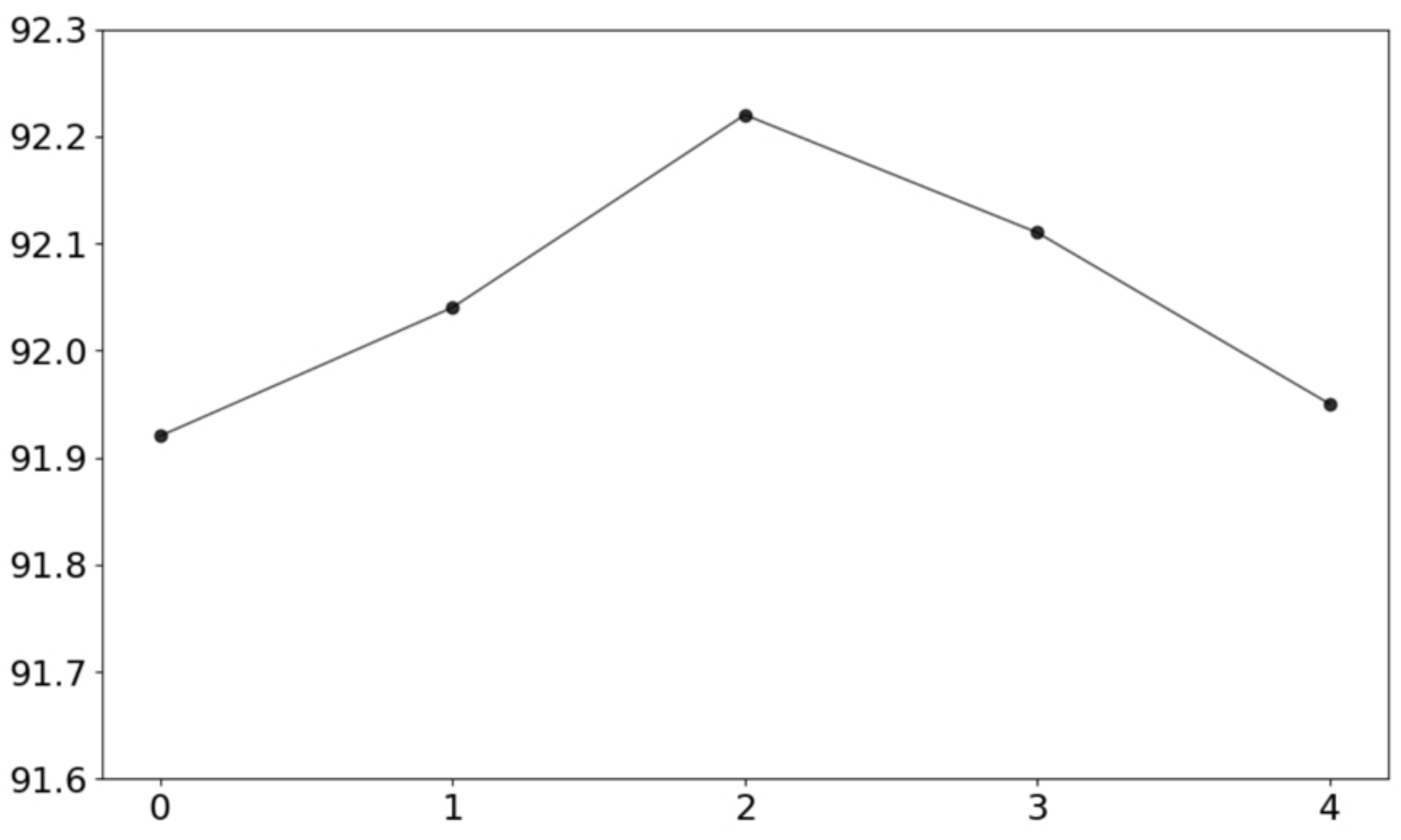}
	\caption{Performance of our model with various window sizes.}
	\label{fig:linechart}
\end{figure}

\begin{table}[!t]
	\small
	\begin{center}
		\begin{tabular}{ c|c|c }
			\hline
			\bf Previous Sentences& \bf Next Sentences\quad & \bf  $F_1$-score$\pm$std \quad \\
			\hline
            $\times$ &$\times$ &91.92$\pm$0.14\\
            \hline
			 \checkmark& $\times$&92.04$\pm$0.13\\
			\hline
             $\times$&\checkmark&91.96$\pm$0.18\\
            \hline
            \checkmark&\checkmark&92.22$\pm$0.02\\
			\hline
		\end{tabular}
	\end{center}
	\caption{Experimental results for ablating additional context in Cross-Sentence Module.}
    \label{tabel_ablation3}
\end{table}

\begin{figure*}
  \centering
	\includegraphics[width=0.98\textwidth]{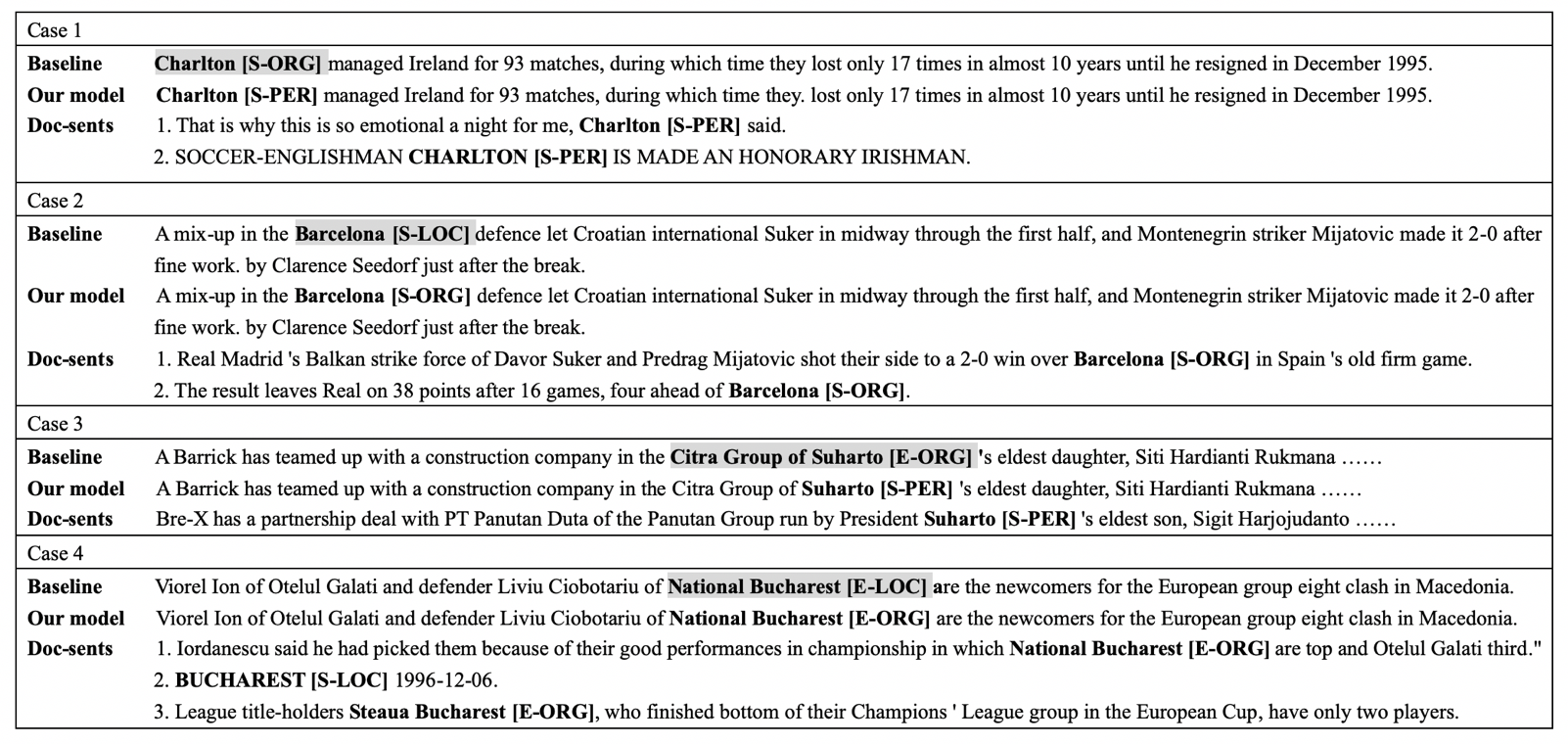}
	\caption{Example of the predictions of Bi-LSTM-CRF baseline and our model.}
	\label{fig:case}
\end{figure*}

\subsubsection{Complexity Analysis}
In this section, we analyze in detail the computation complexity of the proposed GCDoc for a better comparison with the baseline model.
Compared with Bi-LSTM-CRF, GCDoc mainly adds two modules, \ie the cross-sentence module, and the document graph module. Therefore, we first analyze the complexity of these two modules separately. Then we compare the overall complexity of GCDoc and Bi-LSTM-CRF.

\paratitle{The cross-sentence module} includes a sentence encoder based on Bi-LSTM (\rf Eq~\eqref{pre_lstm_1}-\eqref{pre_lstm_3}, Eq~\eqref{method_sentence_encoder}) and an attention mechanism for aggregating cross-sentence context (\rf Eq~\eqref{method_cross_sentence_begin}-\eqref{method_cross_sentence_end}).
The time complexity of the two can be expressed as $O(nd^{2})$ and $O(d^{2})$ respectively, where $n$ denotes the sentence length and $d$ is the dimension of the sentence embeddings.
Therefore, the overall time complexity of the cross-sentence module is $O(nd^{2})$.

\paratitle{The document graph module.}
The computation complexity of this module mainly comes from two parts: one is to use GNN to aggregate the neighbours’ information of each node (\rf Eq~\eqref{edge_weight}, \eqref{final_aggre}), and the other is 
to combine the enhanced representation with the initial embedding via a gating mechanism (\rf Eq~\eqref{eq:ggnn_begin}-\eqref{eq:ggnn_end}).
Note that the size of each node's neighbors varies and some may be very large. Therefore, we sample a fixed-size (\ie $p$) neighborhood of each node as the receptive field during the data preprocessing stage, which can facilitate parallel computing in batches and improve efficiency. In this way, the computational complexity of aggregating the neighbours’ information can be expressed as $O(npd^{2})$. And the gating mechanism for combining representations requires $O(nd^{2})$ computation complexity.
Therefore, the overall time complexity of the document graph module is $O(npd^{2})$.

As for the Bi-LSTM-CRF model, its time complexity can be expressed as $O(nd^{2})$. 
GCDoc adds the above two modules on its basis, thus its overall time complexity is $O(npd^{2})$.
Since $p$ here is actually a constant (set to $5$ in our experiment), the time complexity of GCDoc is also $O(nd^{2})$. Compared with the Bi-LSTM-CRF baseline, our model’s performance on both datasets is significantly improved without leading to a higher computation complexity.

\subsubsection{Impact of Window Size in Cross-Sentence Module}

The window size $k$ (\rf Section \ref{method:cross-sentence}) is clearly a hyperparameter which must be optimized for, thus we investigate the influence of the value of $k$ on the CoNLL2003 NER task. The plot in Figure \ref{fig:linechart} shows that when assigning the value of $k$ to $2$ the model achieves the best results.
When $k$ is less than $2$, the larger $k$ may incorporate more useful contextualized representation by encoding neighboring sentences and improve the results accordingly, when $k$ is larger than $2$, our model drops slightly since more irrelevant context information is involved.



\subsubsection{Case Study}

In this subsection, we present an in-depth analysis of results given by our proposed model. Figure \ref{fig:case} shows four cases that our model predicts correctly but Bi-LSTM-CRF model doesn't. All examples are selected from the CoNLL 2003 test datasets. We also list sentences in the same document that contain words mistakenly tagged by baseline model (\emph{Doc-sents} in the figure) for better understanding the influence of utilizing document-level context in our model.

In the first case, the word ``Charlton" is a person name while baseline recognizes it as an organization (ORG). In the listed \emph{Doc-sents}, PER related contextual information, such as ``said" in the first sentence and ``ENGLISHMAN" in the second sentence indicate ``Charlton" to be a PER. Our model correctly tags ``Charlton" as PER by leveraging document-level contextual information.
In the second case, ``Barcelona" is a polysemous word that can represent a city in spain, or represent an organization as a football club name.
Without obvious ORG related context information, ``Barcelona" is mistakenly labeled as S-LOC by the baseline model.
However, our model successfully labels it as S-ORG by utilizing useful sentences in the document whose context inforamtion strongly indicates ``Barcelona" to be a football club. Similar situation is shown in the third case.
We can infer from ``'s eldest daughter" that the word ``Suharto" represents the name of a person in the sentence, rather than a part of an organization ``City Group of Suharto".
And our model assigns correct label S-PER to it with the help of another sentence in the same document but the baseline model fails.

In the fourth case, we show situation where the document-level context information contains noise. The word ``Bucharest" is part of the orgnazation ``National Bucharest" while baseline mistakenly recognizes it as LOC. 
In the listed \emph{Doc-sents}, in addition to useful contextual information brought by the first and third sentences, ``Bucharest" in the second sentence represents LOC, i.e. a city in Romania, which will inevitibaly introduce noise. 
Our model makes a correct prediction which demenstrates the effectiveness of our model to avoid the interference of noise information.

\subsubsection{Performance on Different Types of Entities}
We further compare the performance of our model and Bi-LSTM-CRF baseline with respect to different types of entities. 
We are interested in the proportion of each type of entity being correctly labeled by different models. Therefore, in Table \ref{analysis_different_entity_type}, we show the corresponding recall scores of the two models with respect to different types of entities on CoNLL03 dataset.

Compared with the Bi-LSTM-CRF model, the recall score of GCDoc drops slightly (0.47\%) on the LOC type, while it's significantly improved on the other three entity types. Among them, for PER and ORG types, GCDoc brings a great improvement of more than 2\%.
We speculate that this is because the document graph module of GCDoc can introduce more useful global information for these two types of entities, because in the CoNLL03 test dataset, there are more scenarios where entity tokens of these two types appear repeatedly in the same document.

\begin{table}[!t]
	\small
	\begin{center}
		\begin{tabular}{m{.13\textwidth}m{.06\textwidth}m{.06\textwidth}m{.06\textwidth}m{.06\textwidth}}
			\toprule
			\bf Model & \bf PER & \bf LOC & \bf ORG & \bf MISC \\
            \midrule
			Bi-LSTM-CRF&95.18& 94.12& 88.44 & 80.91\\
			GCDoc&97.34& 93.65& 90.79& 81.91\\
			$\Delta$ & +2.16 &-0.47 & +2.35 & +1.00 \\
			\bottomrule
		\end{tabular}
	\end{center}
	\caption{Comparison of performance on different type of entities between Bi-LSTM-CRF baseline and our model on CoNLL03 dataset.}
	\label{analysis_different_entity_type}
\end{table}

\section{Conclusions}
This paper proposes a model that exploits \emph{document-level} contextual information for NER at both word-level and sentence-level. 
A document graph is constructed to model wider range of dependencies between words, then obtain an enriched contextual representation via \emph{graph neural networks} (GNN), we further propose 
two strategies to avoid introducing noise information in the document graph module.
A cross-sentence module is also designed to encode adjacent sentences for enriching the contexts of the current sentence. 
Extensive experiments are conducted on two benchmark NER datasets (CoNLL 2003 and Ontonotes 5.0 English datasets) and the results show that our proposed model achieves new state-of-the-art performance.


\bibliography{main}
\bibliographystyle{plain}

\vspace{-1cm}

\end{document}